# Video In Sentences Out

Andrei Barbu,[1,*] Alexander Bridge,[1] Zachary Burchill,[1] Dan Coroian,[1] Sven Dickinson,[2]
Sanja Fidler,[2] Aaron Michaux,[1] Sam Mussman,[1] Siddharth Narayanaswamy,[1] Dhaval Salvi,[3]
Lara Schmidt,[1] Jiangnan Shangguan,[1] Jeffrey Mark Siskind,[1] Jarrell Waggoner,[3] Song Wang,[3]
Jinlian Wei,[1] Yifan Yin,[1] and Zhiqi Zhang[3]

[1]School of Electrical & Computer Engineering, Purdue University, West Lafayette, IN, USA
[2]Department of Computer Science, University of Toronto, Toronto, ON, Canada
[3]Department of Computer Science & Engineering, University of South Carolina, Columbia, SC, USA

## Abstract

We present a system that produces sentential descriptions of video: who did what to whom, and where and how they did it. Action class is rendered as a verb, participant objects as noun phrases, properties of those objects as adjectival modifiers in those noun phrases, spatial relations between those participants as prepositional phrases, and characteristics of the event as prepositional-phrase adjuncts and adverbial modifiers. Extracting the information needed to render these linguistic entities requires an approach to event recognition that recovers object tracks, the track-to-role assignments, and changing body posture.

## 1 INTRODUCTION

We present a system that produces sentential descriptions of short video clips. These sentences describe *who* did *what* to *whom*, and *where* and *how* they did it. This system not only describes the observed action as a verb, it also describes the participant objects as noun phrases, properties of those objects as adjectival modifiers in those noun phrases, the spatial relations between those participants as prepositional phrases, and characteristics of the event as prepositional-phrase adjuncts and adverbial modifiers. It incorporates a vocabulary of 118 words: 1 coordination, 48 verbs, 24 nouns, 20 adjectives, 8 prepositions, 4 lexical prepositional phrases, 4 determiners, 3 particles, 3 pronouns, 2 adverbs, and 1 auxiliary, as illustrated in Table 1.

[*]Corresponding author. Email: andrei@0xab.com.
Additional images and videos as well as all code and datasets are available at
http://engineering.purdue.edu/~qobi/uai2012.

| | |
|---|---|
| **coordination**: | *and* |
| **verbs**: | *approached, arrived, attached, bounced, buried, carried, caught, chased, closed, collided, digging, dropped, entered, exchanged, exited, fell, fled, flew, followed, gave, got, had, handed, hauled, held, hit, jumped, kicked, left, lifted, moved, opened, passed, picked, pushed, put, raised, ran, received, replaced, snatched, stopped, threw, took, touched, turned, walked, went* |
| **nouns**: | *bag, ball, bench, bicycle, box, cage, car, cart, chair, dog, door, ladder, left, mailbox, microwave, motorcycle, object, person, right, skateboard, SUV, table, tripod, truck* |
| **adjectives**: | *big, black, blue, cardboard, crouched, green, narrow, other, pink, prone, red, short, small, tall, teal, toy, upright, white, wide, yellow* |
| **prepositions**: | *above, because, below, from, of, over, to, with* |
| **lexical PPs**: | *downward, leftward, rightward, upward* |
| **determiners**: | *an, some, that, the* |
| **particles**: | *away, down, up* |
| **pronouns**: | *itself, something, themselves* |
| **adverbs**: | *quickly, slowly* |
| **auxiliary**: | *was* |

Table 1: The vocabulary used to generate sentential descriptions of video.

Production of sentential descriptions requires recognizing the primary action being performed, because such actions are rendered as verbs and verbs serve as the central scaffolding for sentences. However, event recognition alone is insufficient to generate the remaining sentential components. One must recognize object classes in order to render nouns. But even object recognition alone is insufficient to generate meaningful sentences. One must determine the *roles* that such objects play in the event. The *agent*, i.e. the doer of the action, is typically rendered as the sentential subject while the *patient*, i.e. the affected object, is typically rendered as the direct object. Detected objects that do not play a role in the observed event, no matter how prominent, should not be incorporated into the description. This means that one cannot use common approaches to event recognition, such as spatiotemporal bags of words [Laptev et al., 2007, Niebles et al., 2008, Scovanner et al., 2007], spatiotemporal volumes [Blank et al., 2005, Laptev et al., 2008, Ro-driguez et al., 2008], and tracked feature points [Liu et al., 2009, Schuldt et al., 2004, Wang and Mori, 2009] that do not determine the class of participant objects and the roles that they play. Even combining such approaches with an object detector would likely detect objects that don't participate in the event and wouldn't be able to determine the roles that any detected objects play.

Producing elaborate sentential descriptions requires more than just event recognition and object detection. Generating a noun phrase with an embedded prepositional phrase, such as *the person to the left of the bicycle*, requires determining spatial relations between detected objects, as well as knowing which of the two detected objects plays a role in the overall event and which serves just to aid generation of a referring expression to help identify the event participant. Generating a noun phrase with adjectival modifiers, such as *the red ball*, not only requires determining the properties, such as color, shape, and size, of the observed objects, but also requires determining whether such descriptions are necessary to help disambiguate the referent of a noun phrase. It would be awkward to generate a noun phrase such as *the big tall wide red toy cardboard trash can* when *the trash can* would suffice. Moreover, one must track the participants to determine the speed and direction of their motion to generate adverbs such as *slowly* and prepositional phrases such as *leftward*. Further, one must track the identity of multiple instances of the same object class to appropriately generate the distinction between *Some person hit some other person* and *The person hit themselves*.

A common assumption in Linguistics [Jackendoff, 1983, Pinker, 1989] is that verbs typically characterize the interaction between event participants in terms of the gross changing motion of these participants. Object class and image characteristics of the participants are believed to be largely irrelevant to determining the appropriate verb label for an action class. Participants simply fill roles in the spatiotemporal structure of the action class described by a verb. For example, an event where one participant (the agent) *picks up* another participant (the patient) consists of a sequence of two sub-events, where during the first sub-event the agent moves towards the patient while the patient is at rest and during the second sub-event the agent moves together with the patient away from the original location of the patient. While determining whether the agent is a *person* or a *cat*, and whether the patient is a *ball* or a *cup*, is necessary to generate the noun phrases incorporated into the sentential description, such information is largely irrelevant to determining the verb describing the action. Similarly, while determining the shapes, sizes, colors, textures, etc. of the participants is necessary to generate adjectival modifiers, such information is also largely irrelevant to determining the verb. Common approaches to event recognition, such as spatiotemporal bags of words, spatiotemporal volumes, and tracked feature points, often achieve high accuracy because of correlation with image or video properties exhibited by a particular corpus. These are often artefactual, not defining properties of the verb meaning (e.g. recognizing *diving* by correlation with *blue* since it 'happens in a pool' [Liu et al., 2009, p. 2002] or confusing *basketball* and *volleyball* 'because most of the time the [...] sports use very similar courts' [Ikizler-Cinibis and Sclaroff, 2010, p. 506]).

## 2 THE MIND'S EYE CORPUS

Many existing video corpora used to evaluate event recognition are ill-suited for evaluating sentential descriptions. For example, the WEIZMANN dataset [Blank et al., 2005] and the KTH dataset [Schuldt et al., 2004] depict events with a single human participant, not ones where people interact with other people or objects. For these datasets, the sentential descriptions would contain no information other than the verb, e.g. *The person jumped*. Moreover, such datasets, as well as the SPORTS ACTIONS dataset [Rodriguez et al., 2008] and the YOUTUBE dataset [Liu et al., 2009], often make action-class distinctions that are irrelevant to the choice of verb, e.g. `wave1` vs. `wave2`, `jump` vs. `pjump`, `Golf-Swing-Back` vs. `Golf-Swing-Front` vs. `Golf-Swing-Side`, `Kicking-Front` vs. `Kicking-Side`, `Swing-Bench` vs. `Swing-SideAngle`, and `golf_swing` vs. `tennis_swing` vs. `swing`. Other datasets, such as the BALLET dataset [Wang and Mori, 2009] and the UCF50 dataset [Liu et al., 2009], depict larger-scale activities that bear activity-class names that are not well suited to sentential description, e.g. `Basketball`, `Billiards`, `BreastStroke`, `CleanAndJerk`, `HorseRace`, `HulaHoop`, `MilitaryParade`, `TaiChi`, and `YoYo`.

The year-one (Y1) corpus produced by DARPA for the Mind's Eye program, however, was specifically designed to evaluate sentential description. This corpus contains two parts: the development corpus, C-D1, which we use solely for training, and the evaluation corpus, C-E1, which we use solely for testing. Each of the above is further divided into four sections to support the four task goals of the Mind's Eye program, namely recognition, description, gap filling, and anomaly detection. In this paper, we use only the recognition and description portions and apply our entire sentential-description pipeline to the combination of these portions. While portions of C-E1 overlap with C-D1, ***in this paper we train our methods solely on C-D1 and test our methods solely on the***

*portion of C-E1 that does not overlap with C-D1*.

Moreover, a portion of the corpus was synthetically generated by a variety of means: computer graphics driven by motion capture, pasting foregrounds extracted from green screening onto different backgrounds, and intensity variation introduced by postprocessing. **In this paper, we exclude all such synthetic video from our test corpus.** Our training set contains 3480 videos and our test set 749 videos. These videos are provided at 720p@30fps and range from 42 to 1727 frames in length, with an average of 435 frames.

The videos nominally depict 48 distinct verbs as listed in Table 1. However, the mapping from videos to verbs is not one-to-one. Due to polysemy, a verb may describe more than one action class, e.g. *leaving an object on the table* vs. *leaving the scene*. Due to synonymy, an action class may be described by more than one verb, e.g. *lift* vs. *raise*. An event described by one verb may contain a component action described by a different verb, e.g. *picking up an object* vs. *touching an object*. Many of the events are described by the combination of a verb with other constituents, e.g. *have a conversation* vs. *have a heart attack*. And many of the videos depict metaphoric extensions of verbs, e.g. *take a puff on a cigarette*. Because the mapping from videos to verbs is subjective, the corpus comes labeled with DARPA-collected human judgments in the form of a single present/absent label associated with each video paired with each of the 48 verbs, gathered using Amazon Mechanical Turk. We use these labels for both training and testing as described later.

## 3   OVERALL SYSTEM ARCHITECTURE

The overall architecture of our system is depicted in Fig. 1. We first apply detectors [Felzenszwalb et al., 2010a,b] for each object class on each frame of each video. These detectors are biased to yield many false positives but few false negatives. The Kanade-Lucas-Tomasi (KLT) [Shi and Tomasi, 1994, Tomasi and Kanade, 1991] feature tracker is then used to project each detection five frames forward to augment the set of detections and further compensate for false negatives in the raw detector output. A dynamic-programming algorithm [Viterbi, 1971] is then used to select an optimal set of detections that is temporally coherent with optical flow, yielding a set of object tracks for each video. These tracks are then smoothed and used to compute a time-series of feature vectors for each video to describe the relative and absolute motion of event participants. The person detections are then clustered based on part displacements to derive a coarse measure of human body posture in the form of a body-posture codebook. The codebook indices of person detections are then added to the feature vector. Hidden Markov Models (HMMs) are then employed as time-series classifiers to yield verb labels for each video [Siskind and Morris, 1996, Starner et al., 1998, Wang and Mori, 2009, Xu et al., 2002, 2005], together with the object tracks of the participants in the action described by that verb along with the roles they play. These tracks are then processed to produce nouns from object classes, adjectives from object properties, prepositional phrases from spatial relations, and adverbs and prepositional-phrase adjuncts from track properties. Together with the verbs, these are then woven into grammatical sentences. We describe each of the components of this system in detail below: the object detector and tracker in Section 3.1, the body-posture clustering and codebook in Section 3.2, the event classifier in Section 3.3, and the sentential-description component in Section 3.4.

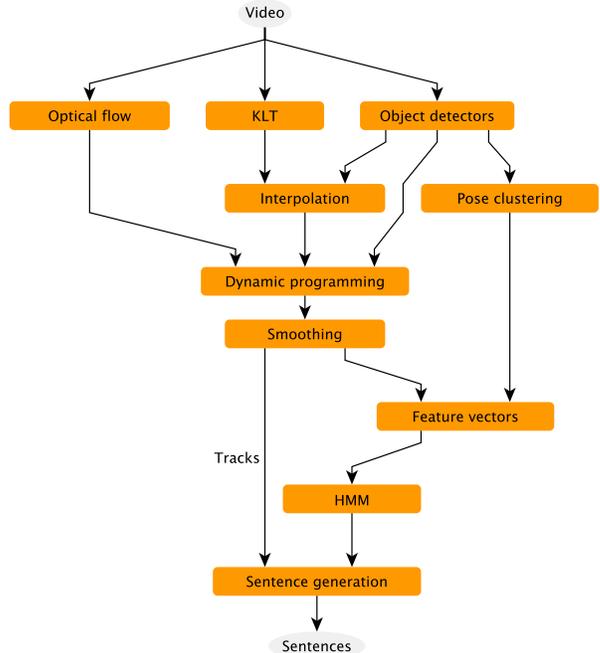

Figure 1: The overall architecture of our system for producing sentential descriptions of video.

### 3.1   OBJECT DETECTION AND TRACKING

In detection-based tracking an object detector is applied to each frame of a video to yield a set of candidate detections which are composed into tracks by selecting a single candidate detection from each frame that maximizes temporal coherency of the track. Felzenszwalb et al. detectors are used for this purpose. Detection-based tracking requires biasing the detector to have high recall at the expense of low precision to allow the tracker to select boxes to yield a temporally coherent track. This is done by depressing the acceptance thresholds. To prevent massive over-generation of false positives, which would severely impact run time, we limit the number of detections produced per-frame to 12.

Two practical issues arise when depressing acceptance thresholds. First, it is necessary to reduce the degree of non-maximal suppression incorporated in the Felzenszwalb et al. detectors. Second, with the star detector [Felzenszwalb et al., 2010b], one can simply decrease the single trained acceptance threshold to yield more detections with no increase in computational complexity. However, we prefer to use the star cascade detector [Felzenszwalb et al., 2010a] as it is far faster. With the star cascade detector, though, one must also decrease the trained root- and part-filter thresholds to get more detections. Doing so, however, defeats the computational advantage of the cascade and significantly increases detection time. We thus train a model for the star detector using the standard procedure on human-annotated training data, sample the top detections produced by this model with a decreased acceptance threshold, and train a model for the star cascade detector on these samples. This yields a model that is almost as fast as one trained by the star cascade detector on the original training samples but with the desired bias in acceptance threshold.

The Y1 corpus contains approximately 70 different object classes that play a role in the depicted events. Many of these, however, cannot be reliably detected with the Felzenszwalb et al. detectors that we use. We trained models for 25 object classes that can be reliably detected, as listed in Table 2. These object classes account for over 90% of the event participants. Person models were trained with approximately 2000 human-annotated positive samples from C-D1 while nonperson models were trained with approximately 1000 such samples. For each positive training sample, two negative training samples were randomly generated from the same frame constrained to not overlap substantially with the positive samples. We trained three distinct person models to account for body-posture variation and pool these when constructing person tracks. The detection scores were normalized for such pooled detections by a per-model offset computed as follows: A (50 bin) histogram was computed of the scores of the top detection in each frame of a video. The offset is then taken to be the minimum of the value that maximizes the between-class variance [Otsu, 1979] when bipartitioning this histogram and the trained acceptance threshold offset by a fixed, but small, amount (0.4).

We employed detection-based tracking for all 25 object models on all 749 videos in our test set. To prune the large number of tracks thus produced, we discard all tracks corresponding to certain object models on a per-video basis: those that exhibit high detection-score variance over the frames in that video as well as those whose detection-score distributions are neither unimodal nor bimodal. The parameters governing such pruning were determined solely on the training set. The tracks that remain after this pruning still account for over 90% of the event participants.

### 3.2 BODY-POSTURE CODEBOOK

We recognize events using a combination of the motion of the event participants and the changing body posture of the human participants. Body-posture information is derived using the part structure produced as a by-product of the Felzenszwalb et al. detectors. While such information is far noisier and less accurate than fitting precise articulated models [Andriluka et al., 2008, Bregler, 1997, Gavrila and Davis, 1995, Sigal et al., 2010, Yang and Ramanan, 2011] and appears unintelligible to the human eye, as shown in Section 3.3, it suffices to improve event-recognition accuracy. Such information can be extracted from a large unannotated corpus far more robustly than possible with precise articulated models.

Body-posture information is derived from part structure in two ways. First, we compute a vector of part displacements, each displacement as a vector from the detection center to the part center, normalizing these vectors to unit detection-box area. The time-series of feature vectors is augmented to include these part displacements and a finite-difference approximation of their temporal derivatives as continuous features for person detections. Second, we vector-quantize the part-displacement vector and include the codebook index as a discrete feature for person detections. Such pose features are included in the time-series on a per-frame basis. The codebook is trained by running each pose-specific person detector on the positive human-annotated samples used to train that detector and extract the resulting part-displacement vectors. We then pool the part-displacement vectors from the three pose-specific person models and employ hierarchical $k$-means clustering using Euclidean distance to derive a codebook of 49 clusters. Fig. 2 shows sample clusters from our codebook. Codebook indices are derived using Euclidean distance from the means of these clusters.

|     |                                                                                                                                                                                                 |
| --- | ----------------------------------------------------------------------------------------------------------------------------------------------------------------------------------------------- |
| (a) | bag↦*bag*  bench↦*bench*  bicycle↦*bicycle*  big-ball↦*ball*  cage↦*cage*  car↦*car*  cardboard-box↦*box*  cart↦*cart*  chair↦*chair*  dog↦*dog*  door↦*door*  ladder↦*ladder*  mailbox↦*mailbox*  microwave↦*microwave*  motorcycle↦*motorcycle*  person↦*person*  person-crouch↦*person*  person-down↦*person*  skateboard↦*skateboard*  small-ball↦*ball*  suv↦*SUV*  table↦*table*  toy-truck↦*truck*  tripod↦*tripod*  truck↦*truck* |
| (b) | cardboard-box↦*cardboard*   person↦*upright*   person-crouch↦*crouched*   person-down↦*prone*   toy-truck↦*toy*                                                                                 |
| (c) | big-ball↦*big*   small-ball↦*small*                                                                                                                                                              |

Table 2: Trained models for object classes and their mappings to (a) nouns, (b) restrictive adjectives, and (c) size adjectives.

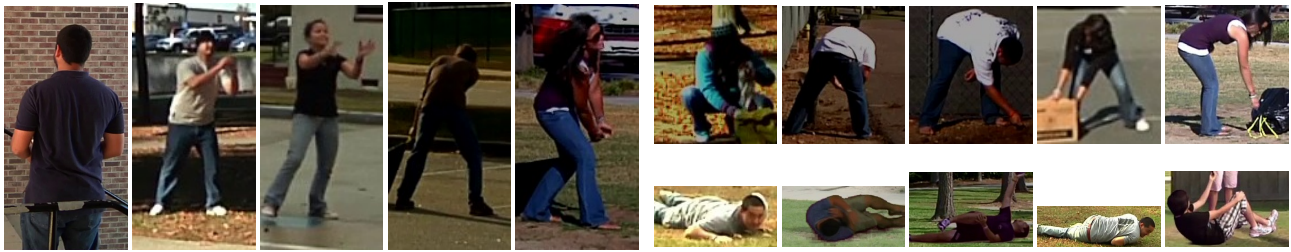

Figure 2: Sample clusters from our body-posture codebook.

### 3.3   EVENT CLASSIFICATION

Our tracker produces one or more tracks per object class for each video. We convert such tracks into a time-series of feature vectors. For each video, one track is taken to designate the agent and another track (if present) is taken to designate the patient. During training, we manually specify the track-to-role mapping. During testing, we automatically determine the track-to-role mapping by examining all possible such mappings and selecting the one with the highest likelihood [Siskind and Morris, 1996].

The feature vector encodes both the motion of the event participants and the changing body posture of the human participants. For each event participant in isolation we incorporate the following single-track features:

1. $x$ and $y$ coordinates of the detection-box center
2. detection-box aspect ratio and its temporal derivative
3. magnitude and direction of the velocity of the detection-box center
4. magnitude and direction of the acceleration of the detection-box center
5. normalized part displacements and their temporal derivatives
6. object class (the object detector yielding the detection)
7. root-filter index
8. body-posture codebook index

The last three features are discrete; the remainder are continuous. For each pair of event participants we incorporate the following track-pair features:

1. distance between the agent and patient detection-box centers and its temporal derivative
2. orientation of the vector from agent detection-box center to patient detection-box center

Our HMMs assume independent output distributions for each feature. Discrete features are modeled with discrete output distributions. Continuous features denoting linear quantities are modeled with univariate Gaussian output distributions, while those denoting angular quantities are modeled with von Mises output distributions.

For each of the 48 action classes, we train two HMMs on two different sets of time-series of feature vectors, one containing only single-track features for a single participant and the other containing single-track features for two participants along with the track-pair features. A training set of between 16 and 200 videos was selected manually from C-D1 for each of these 96 HMMs as positive examples depicting each of the 48 action classes. A given video could potentially be included in the training sets for both the one-track and two-track HMMs for the same action class and even for HMMs for different action classes, if the video was deemed to depict both action classes.

During testing, we generate present/absent judgments for each video in the test set paired with each of the 48 action classes. We do this by thresholding the likelihoods produced by the HMMs. By varying these thresholds, we can produce an ROC curve for each action class, comparing the resulting machine-generated present/absent judgments with the Amazon Mechanical Turk judgments. When doing so, we test videos for which our tracker produces two or more tracks against only the two-track HMMs while we test ones for which our tracker produces a single track against only the one-track HMMs.

We performed three experiments, training 96 different 200-state HMMs for each. Experiment I omitted all discrete features and all body-posture related features. Experiment II omitted only the discrete features. Experiment III omitted only the continuous body-posture related features. ROC curves for each experiment are shown in Fig. 3. Note that the incorporation of body-posture information, either in the form of continuous normalized part displacements or discrete codebook indices, improves event-recognition accuracy, despite the fact that the part displacements produced by the Felzenszwalb et al. detectors are noisy and appear unintelligible to the human eye.

### 3.4 GENERATING SENTENCES

We produce a sentence from a detected action class together with the associated tracks using the templates from Table 3. In these templates, words in *italics* denote fixed strings, words in **bold** indicate the action class, X and Y denote subject and object noun phrases, and the categories Adv, PP$_{\text{endo}}$, and PP$_{\text{exo}}$ denote adverbs and prepositional-phrase adjuncts to describe the subject motion. The processes for generating these noun phrases, adverbs, and prepositional-phrase adjuncts are described below. One-track HMMs take that track to be the agent and thus the subject. For two-track HMMs we choose the mapping from tracks to roles that yields the higher likelihood and take the agent track to be the subject and the patient track to be the object except when the action class is either **approached** or **fled**, the agent is (mostly) stationary, and the patient moves more than the agent.

Brackets in the templates denote optional entities. Optional entities containing Y are generated only for two-track HMMs. The criteria for generating optional adverbs and prepositional phrases are described below. The optional entity for **received** is generated when there is a patient track whose category is `mailbox`, `person`, `person-crouch`, or `person-down`.

We use adverbs to describe the velocity of the subject. For some verbs, a velocity adverb would be awkward:

∗X ***slowly*** *had* Y      ∗X *had **slowly*** Y

Furthermore, stylistic considerations dictate the syntactic position of an optional adverb:

X *jumped **slowly** over* Y    X ***slowly*** *jumped over* Y
X ***slowly** approached* Y     ∗X *approached **slowly*** Y
?X ***slowly** fell*            X *fell **slowly***

The verb-phrase templates thus indicate whether an adverb is allowed, and if so whether it occurs, preferentially, preverbally or postverbally. Adverbs are chosen subject to three thresholds $v_1^{\text{action class}}$, $v_2^{\text{action class}}$, and $v_3^{\text{action class}}$ determined empirically on a per-action-class basis: We select those frames from the subject track where the magnitude of the velocity of the box-detection center is above $v_1^{\text{action class}}$. An optional adverb is generated by comparing the magnitude of the average velocity $v$ of the subject track box-detection centers in these frames to the per-action-class thresholds:

$$\begin{aligned} quickly & \quad v > v_2^{\text{action class}} \\ slowly & \quad v_1^{\text{action class}} \leq v \leq v_3^{\text{action class}} \end{aligned}$$

We use prepositional-phrase adjuncts to describe the motion direction of the subject. Again, for some verbs, such adjuncts would be awkward:

∗X *had* Y ***leftward***      ∗X *had* Y ***from the left***

Moreover, for some verbs it is natural to describe the motion direction endogenously, from the perspective of the subject, while for others it is more natural to describe the motion direction exogenously, from the perspective of the viewer:

X *fell **leftward***           X *fell **from the left***
X *chased* Y ***leftward***     ∗X *chased* Y ***from the left***
∗X *arrived **leftward***       X *arrived **from the left***

The verb-phrase templates thus indicate whether an adjunct is allowed, and if so whether it is preferentially endogenous or exogenous. The choice of adjunct is determined from the orientation of $v$, as computed above and depicted in Fig. 4(a,b). We omit the adjunct when $v < v_1^{\text{action class}}$.

We generate noun phrases X and Y to refer to event participants according to the following grammar:

NP → *themselves* | *itself* | *something* | D A* N [PP]
D  → *the* | *that* | *some*

When instantiating a sentential template that has a required object noun-phrase Y for a one-track HMM, we generate a pronoun. A pronoun is also generated when the action class is **entered** or **exited** and the patient class is not `car`, `door`, `suv`, or `truck`. The anaphor *themselves* is generated if the action class is **attached** or **raised**, the anaphor *itself* if the action class is **moved**, and *something* otherwise.

As described below, we generate an optional prepositional phrase for the subject noun phrase to describe the spatial relation between the subject and the object. We choose the determiner to handle coreference, generating *the* when a noun phrase unambiguously refers to the agent or the patient due to the combination of head noun and any adjectives,

***The*** *person jumped over **the** ball.*
***The*** *red ball collided with **the** blue ball.*

*that* for an object noun phrase that corefers to a track referred to in a prepositional phrase for the subject,

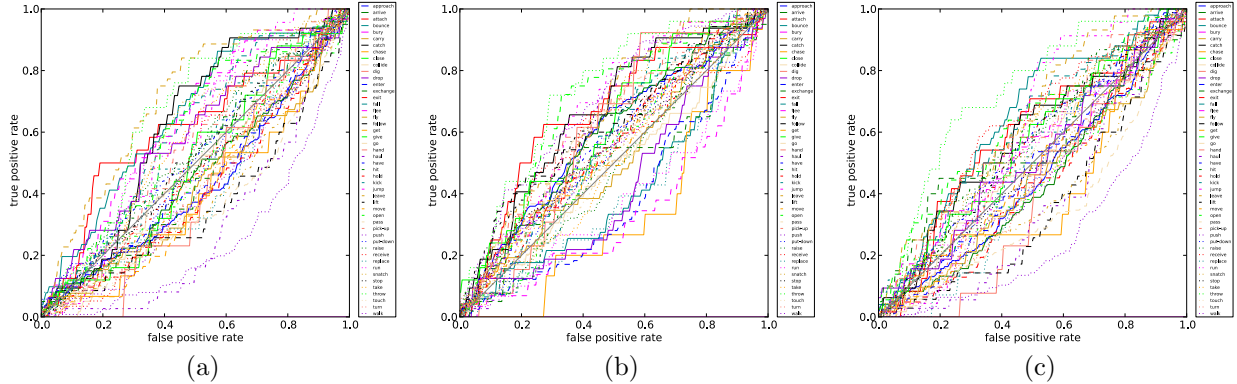

Figure 3: ROC curves for each of the 48 action classes for (a) Experiment I, omitting all discrete and body-posture-related features, (b) Experiment II, omitting only the discrete features, and (c) Experiment III, omitting only the continuous body-posture-related features.

| | | | |
|---|---|---|---|
| X [Adv] **approached** Y [PP$_{exo}$] | X [Adv] **entered** Y [PP$_{endo}$] | X **had** Y | X **put** Y **down** |
| X **arrived** [Adv] [PP$_{exo}$] | X [Adv] **exchanged** *an object with* Y | X **hit** [*something with*] Y | X **raised** Y |
| X [Adv] **attached** *an object to* Y | X [Adv] **exited** Y [PP$_{endo}$] | X **held** Y | X **received** [*an object from*]Y |
| X **bounced** [Adv] [PP$_{endo}$] | X **fell** [Adv] [*because of* Y] [PP$_{endo}$] | X **jumped** [Adv] [*over* Y] [PP$_{endo}$] | X [Adv] **replaced** Y |
| X **buried** Y | X **fled** [Adv] [*from* Y] [PP$_{endo}$] | X [Adv] **kicked** Y [PP$_{endo}$] | X **ran** [Adv] [*to* Y] [PP$_{endo}$] |
| X [Adv] **carried** Y [PP$_{endo}$] | X **flew** [Adv] [PP$_{endo}$] | X [Adv] **left** Y [PP$_{endo}$] | X [Adv] **snatched** *an object from* Y |
| X **caught** Y [PP$_{exo}$] | X [Adv] **followed** Y [PP$_{endo}$] | X [Adv] **lifted** Y | X [Adv] **stopped** [Y] |
| X [Adv] **chased** Y [PP$_{endo}$] | X **got** *an object from* Y | X [Adv] **moved** Y [PP$_{endo}$] | X [Adv] **took** *an object from* Y |
| X **closed** Y | X **gave** *an object to* Y | X **opened** Y | X [Adv] **threw** Y [PP$_{endo}$] |
| X [Adv] **collided** *with* Y [PP$_{exo}$] | X **went** [Adv] *away* [PP$_{endo}$] | X [Adv] **passed** Y [PP$_{exo}$] | X **touched** Y |
| X *was* **digging** [*with* Y] | X **handed** Y *an object* | X **picked** Y **up** | X **turned** [PP$_{endo}$] |
| X **dropped** Y | X [Adv] **hauled** Y [PP$_{endo}$] | X [Adv] **pushed** Y [PP$_{endo}$] | X **walked** [Adv] [*to* Y] [PP$_{endo}$] |

Table 3: Sentential templates for the action classes indicated in bold.

**The** *person to the right of* **the** *car approached* **that** *car.*
**Some** *person to the right of* **some** *other person approached* **that** *other person.*

and *some* otherwise:

**Some** *car approached* **some** *other car.*

We generate the head noun of a noun phrase from the object class using the mapping in Table 2(a). Four different kinds of adjectives are generated: color, shape, size, and restrictive modifiers. An optional color adjective is generated based on the average HSV values in the eroded detection boxes for a track: *black* when $V \leq 0.2$, *white* when $V \geq 0.8$, one of *red*, *blue*, *green*, *yellow*, *teal*, or *pink* based on $H$, when $S \geq 0.7$. An optional size adjective is generated in two ways, one from the object class using the mapping in Table 2(c), the other based on per-object-class image statistics. For each object class, a mean object size $\bar{a}_{\text{object class}}$ is determined by averaging the detected-box areas over all tracks for that object class in the training set used to train HMMs. An optional size adjective for a track is generated by comparing the average detected-box area $a$ for that track to $\bar{a}_{\text{object class}}$:

*big*    $a \geq \beta_{\text{object class}}\bar{a}_{\text{object class}}$
*small*    $a \leq \alpha_{\text{object class}}\bar{a}_{\text{object class}}$

The per-object-class cutoff ratios $\alpha_{\text{object class}}$ and $\beta_{\text{object class}}$ are computed to equally tripartition the distribution of per-object-class mean object sizes on the training set. Optional shape adjectives are generated in a similar fashion. Per-object-class mean aspect ratios $\bar{r}_{\text{object class}}$ are determined in addition to the per-object-class mean object sizes $\bar{a}_{\text{object class}}$. Optional shape adjectives for a track are generated by comparing the average detected-box aspect ratio $r$ and area $a$ for that track to these means:

*tall*    $r \leq 0.7\bar{r}_{\text{object class}} \wedge a \geq \beta_{\text{object class}}\bar{a}_{\text{object class}}$
*short*    $r \geq 1.3\bar{r}_{\text{object class}} \wedge a \leq \alpha_{\text{object class}}\bar{a}_{\text{object class}}$
*narrow*    $r \leq 0.7\bar{r}_{\text{object class}} \wedge a \leq \alpha_{\text{object class}}\bar{a}_{\text{object class}}$
*wide*    $r \geq 1.3\bar{r}_{\text{object class}} \wedge a \geq \beta_{\text{object class}}\bar{a}_{\text{object class}}$

To avoid generating shape and size adjectives for unstable tracks, they are only generated when the detection-score variance and the detected aspect-ratio variance for the track are below specified thresholds. Optional restrictive modifiers are generated from the object class using the mapping in Table 2(b). Person-pose adjectives are generated from aggregate body-posture information for the track: object class, normalized part displacements, and body-posture codebook indices. We generate all applicable adjectives except for color and person pose. Following the Gricean Maxim of Quantity [Grice, 1975], we only generate color and person-pose adjectives if needed to prevent coreference of nonhuman event participants. Finally, we generate an initial adjective *other*, as needed to prevent coreference. Generating *other* does not allow generation of the determiner *the* in place of *that* or *some*. We order any adjectives generated so that *other* comes first, followed by size, shape, color, and restrictive modifiers, in that order.

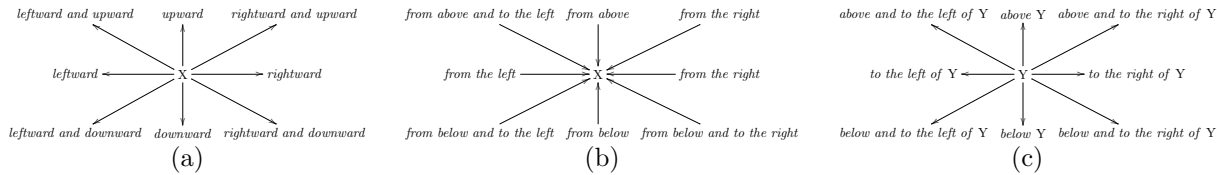

Figure 4: (a) Endogenous and (b) exogenous prepositional-phrase adjuncts to describe subject motion direction. (c) Prepositional phrases incorporated into subject noun phrases describing viewer-relative 2D spatial relations between the subject X and the reference object Y.

For two-track HMMs where neither participant moves, a prepositional phrase is generated for subject noun phrases to describe the static 2D spatial relation between the subject X and the reference object Y from the perspective of the viewer, as shown in Fig. 4(c).

## 4  EXPERIMENTAL RESULTS

We used the HMMs generated for Experiment III to compute likelihoods for each video in our test set paired with each of the 48 action classes. For each video, we generated sentences corresponding to the three most-likely action classes. Fig. 5 shows key frames from four videos in our test set along with the sentence generated for the most-likely action class. Human judges rated each video-sentence pair to assess whether the sentence was true of the video and whether it described a salient event depicted in that video. 26.7% (601/2247) of the video-sentence pairs were deemed to be true and 7.9% (178/2247) of the video-sentence pairs were deemed to be salient. When restricting consideration to only the sentence corresponding to the single most-likely action class for each video, 25.5% (191/749) of the video-sentence pairs were deemed to be true and 8.4% (63/749) of the video-sentence pairs were deemed to be salient. Finally, for 49.4% (370/749) of the videos at least one of the three generated sentences was deemed true and for 18.4% (138/749) of the videos at least one of the three generated sentences was deemed salient.

## 5  CONCLUSION

Integration of Language and Vision [Aloimonos et al., 2011, Barzialy et al., 2003, Darrell et al., 2011, McKevitt, 1994, 1995–1996] and recognition of action in video [Blank et al., 2005, Laptev et al., 2008, Liu et al., 2009, Rodriguez et al., 2008, Schuldt et al., 2004, Siskind and Morris, 1996, Starner et al., 1998, Wang and Mori, 2009, Xu et al., 2002, 2005] have been of considerable interest for a long time. There has also been work on generating sentential descriptions of static images [Farhadi et al., 2009, Kulkarni et al., 2011, Yao et al., 2010]. Yet we are unaware of any prior work that generates as rich sentential video descriptions as we describe here. Producing such rich descriptions requires determining event participants, the mapping of such participants to roles in the event, and their motion and properties. This is incompatible with common approaches to event recognition, such as spatiotemporal bags of words, spatiotemporal volumes, and tracked feature points that cannot determine such information. The approach presented here recovers the information needed to generate rich sentential descriptions by using detection-based tracking and a body-posture codebook. We demonstrated the efficacy of this approach on a corpus of 749 videos.

**Acknowledgments**

This work was supported, in part, by NSF grant CCF-0438806, by the Naval Research Laboratory under Contract Number N00173-10-1-G023, by the Army Research Laboratory accomplished under Cooperative Agreement Number W911NF-10-2-0060, and by computational resources provided by Information Technology at Purdue through its Rosen Center for Advanced Computing. Any views, opinions, findings, conclusions, or recommendations contained or expressed in this document or material are those of the author(s) and do not necessarily reflect or represent the views or official policies, either expressed or implied, of NSF, the Naval Research Laboratory, the Office of Naval Research, the Army Research Laboratory, or the U.S. Government. The U.S. Government is authorized to reproduce and distribute reprints for Government purposes, notwithstanding any copyright notation herein.

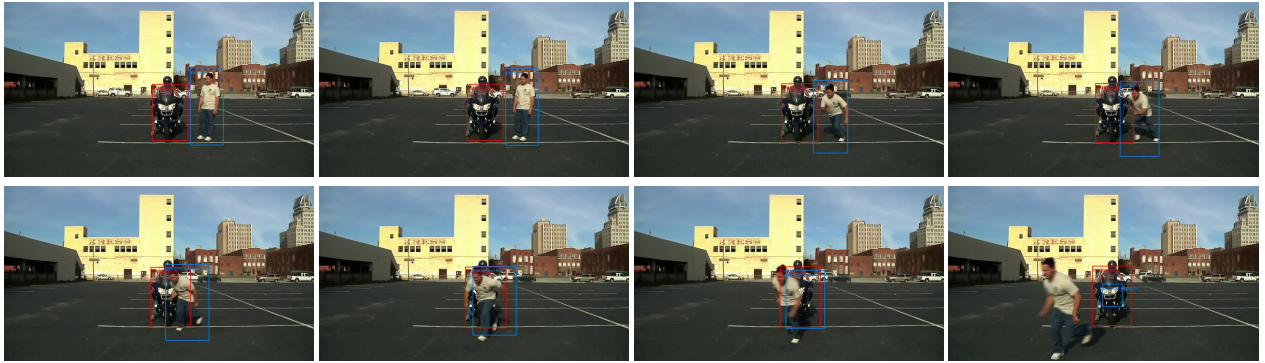

The upright person to the right of the motorcycle went away leftward.

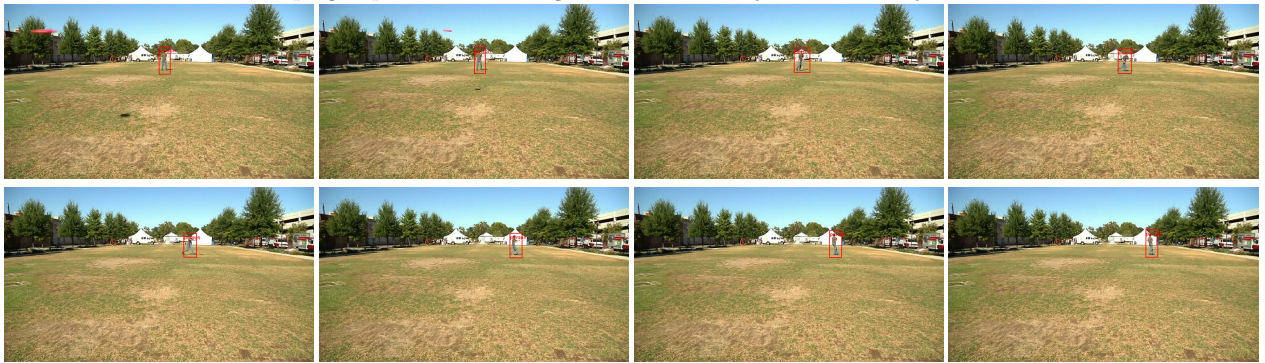

The person walked slowly to something rightward.

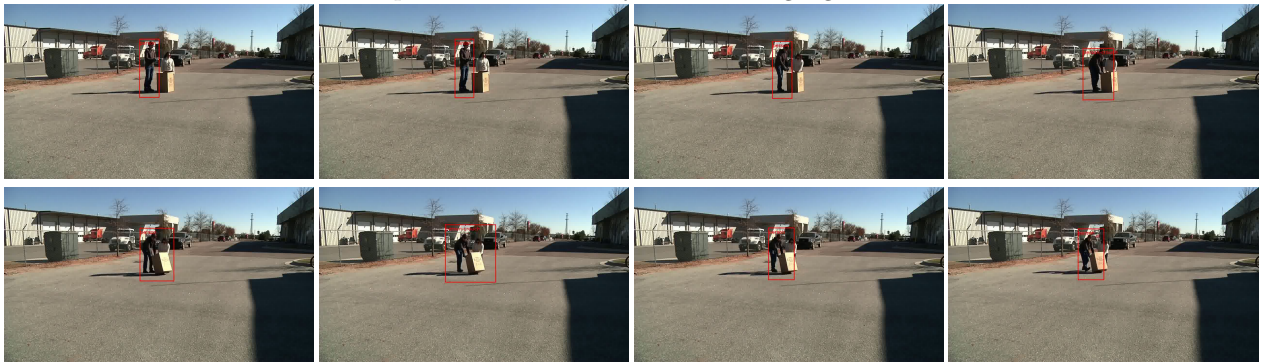

The narrow person snatched an object from something.

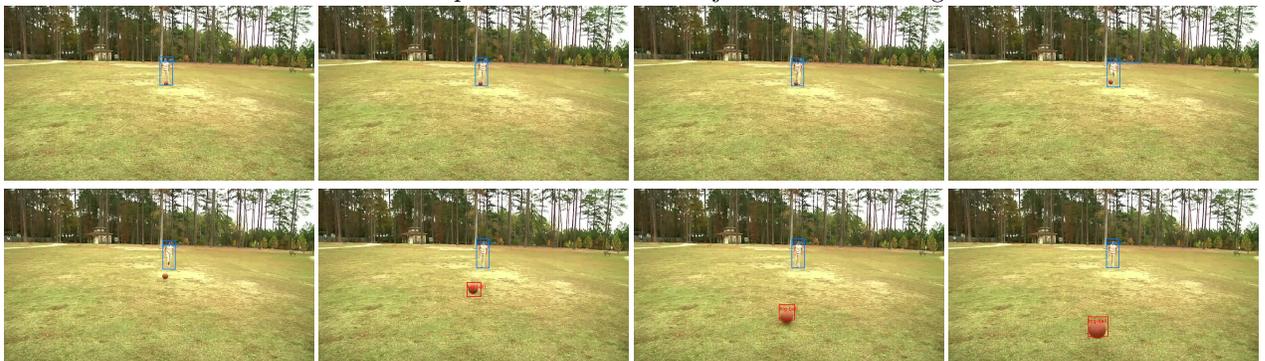

The upright person hit the big ball.

Figure 5: Key frames from four videos in our test set along with the sentence generated for the most-likely action class.